\documentclass[letterpaper]{article}
\usepackage{iccc}

\usepackage{float}
\usepackage{url}
\usepackage{latexsym}
\usepackage{subcaption}
\usepackage{graphicx}
\usepackage{nameref}
\usepackage[czech,english]{babel}
\usepackage{booktabs}

\usepackage{times}
\usepackage{helvet}
\usepackage{courier}
\title{GPT Czech Poet: Generation of Czech Poetic Strophes with Language Models}
\author{Michal Chudoba \and Rudolf Rosa\\
Institute of Formal and Applied Linguistics, Faculty of Mathematics and Physics\\
Charles University, Praha, Czech Republic\\
michal.chudoba.praha@gmail.com \and rosa@ufal.mff.cuni.cz }
\begin{document} 
\maketitle
\begin{abstract}
\begin{quote}
High-quality automated poetry generation systems are currently only available for a small subset of languages.
We introduce a new model for generating poetry in Czech language, based on fine-tuning a pre-trained Large Language Model.
We demonstrate that guiding the generation process by explicitly specifying strophe parameters within the poem text strongly improves the effectiveness of the model.
We also find that appropriate tokenization is crucial, showing that tokenization methods based on syllables or individual characters instead of subwords prove superior in generating poetic strophes.
We further enhance the results by introducing \textit{Forced~generation}, adding explicit specifications of meter and verse parameters at inference time based on the already generated text.
We evaluate a range of setups, showing that our proposed approach achieves high accuracies in rhyming and metric aspects of formal quality of the generated poems.
\end{quote}
\end{abstract}

\section{Introduction}

In this work, our goal is to devise a generative language model to generate poem strophes in Czech language, according to specified rhyme scheme and meter.

With the arrival of pre-trained open-domain Large Language Models (LLMs), such as GPT \cite{Radford2019LanguageMA,openai2023gpt4} or Llama \cite{touvron2023llama}, the two most common approaches for solving such tasks is either fine-tuning a LLM on a specialized dataset, or prompting the LLM without training.
The prompting approach is applicable for languages and tasks sufficiently prevalent in the training data \cite{Liu_2023}, and might thus be applicable for English poetry to some extent.

Our preliminary experiments have shown that even the best available models, such as GPT-4, struggle to adhere to the structural nuances of strophes and associated parameters of Czech poetry, leading to low-quality outputs.
Therefore, we adopt the approach of fine-tuning an available LLM, specifically Czech GPT-2 \texttt{czech-gpt2-oscar} by Chaloupský~\shortcite{medicalczgpt}, on a large corpus of Czech poetry.
The Czech language also differs in several important characteristics from languages typically studied in previous works, most notably by its rich inflection but rather regular orthography and prosody, which motivates the approach we take in this work.

%


We draw inspiration from treating text as a sequence of syllables~\cite{oncevay2020revisiting}. 
Our main emphasis is not on the detailed meaning of the text, an area where models using standard tokenizers like BPE~\cite{bpe} perform well. Instead, we prioritize the phonetic aspects and adherence to meter, which are crucial for our task. Syllabic modeling proves particularly advantageous in generating neologisms, common in poetry to maintain prescribed rhyme scheme and meter. In pursuit of this, we have delved into tokenizer-free models~\cite{xue2022byt5}, offering maximal flexibility in constructing neologisms and pairing characters to align with stipulated strophe parameters. This approach, already demonstrated to be effective in poetry generation by the ByGPT5 system~\cite{belouadi2023bygpt5}, showcased proficiency in both rhyme scheme and meter adherence. We also experiment with several ways of guiding the generation process by interleaving explicit annotations with the strophe text.

\begin{figure}[H]
\begin{tabular}{|lll}
    Tvá loď jde po vy-so-kém mo-ři, & \textbf{A} & \emph{iamb}\\
    x~~~~~X~~~~~x~~~~X~~~x~~~x~~~~x~~~~X~~~x & & \\
    v ně brá-zdu ja-ko stří-bro re-je, & \textbf{B} & \emph{iamb}\\
    x~~~~~X~~~~~x~~~X~~~x~~~~X~~~~x~~~X~~~x & & \\
    svou pří-du v mod-ré vl-ny no-ří  & \textbf{A} & \emph{iamb}\\
    x~~~~~~~X~~~~x~~~~~X~~~~x~~~X~~x~~~X~~~x & & \\
    a bok svůj pěn-né do pe-ře-je.  & \textbf{B} & \emph{iamb}\\
    x~~~X~~~~~x~~~~X~~~~x~~~X~~x~~~x~~~x & &
\end{tabular}
    \caption{An \textbf{ABAB} strophe with meter annotation and rhythm annotation: `x` = unstressed syllable, `X` = stressed syllable. (\textit{Your ship is on the high seas, with a furrow in it like silver, she plunges her prow into the blue waves, and its side foaming into the rapids.})} 
    \label{fig:example-strophe-annotation}
\end{figure}

\subsection{Contributions}
The main contributions of this paper are:
\begin{itemize}
    \item advancing poetry generation in Czech language,
    \item introduction of providing verse parameter hints (number of syllables, verse ending) to encourage rhyming,
    \item evaluation of several schemes of encoding rhyme, meter and style,
    \item evaluation of several tokenization schemes.
\end{itemize}

\section{Parameters of Poetry}
In poetic strophes, there are two main parameters that govern their structure: rhyme and meter (even though many strophes are crafted without adhering to rhyme or are constructed in free verse). While the rhyme scheme applies to the entire strophe, the meter may vary from verse to verse. Consequently, in our analysis, we meticulously annotate the meter for each individual verse (see example in Figure~\ref{fig:example-strophe-annotation}).

\subsection{Rhyme} Utilizing the standard approach, we designate the rhyme scheme with capital letters, such as \textbf{ABAB}, where each character denotes an individual verse in the strophe, also allowing \textbf{X} for non-rhyming verses.
We include configurations of both 4 and 6 lines.
The rhyming scheme thus can be e.g.\ \textbf{AABBCC}, where each verse has a corresponding rhyming pair,
as well as e.g.\ \textbf{XAXA}, where only the second verse rhymes with the fourth.

\subsection{Meter} We considered the following meter types that occur in our dataset (labelled with one-letter labels), according to Plecháč et al.~\shortcite{ccv2015}:
\begin{description}
    \item[iamb (J)] 
    First syllable is  unstressed, second is stressed.
    E.g.\ 
    `attempt' to `at-tempt', stress is on second syllable `tempt'.
    \item[trochee (T)]
    Reverse of iamb, first syllable is stressed, second is unstressed.
    E.g.\ `double' to `dou-ble' with stress on first syllable.
    \item[dactyl (D)]
    Three part meter with stress on first syllable. Next two syllables are unstressed. E.g.\ `poetry' to `po-et-ry' with stress on first syllable.
    \item[amphibrach (A)]
    Three part meter with stress on second syllable. E.g.\ `the scenes of', where stress is placed on the word `scenes'.
    \item[dactylotrochee (X)]
    Combination of dactyl and trochee.
    \item[dactylotrochee with anacrusis (Y)]
    Anacrusis is a set of unstressed syllables preceding the first stressed dactylotrochee syllable.
    \item[hexameter (H)]
    Dactyl meter of 6 parts, where 5th part must be dactyl and 6th trochee.
    \item[pentameter (P)]
    Dactyl meter of 5 parts, where 3rd and 6th are not complete.
    \item[Not Recognized (N)]
    Does not pertain to any before mentioned meter. Consists of free-verse, Syllable weight based meter and not recognized meters.
\end{description}

See Figure \ref{fig:example-strophe-annotation} for an example of a strophe with the \textbf{ABAB} rhyme scheme and \textit{iamb} meter for each verse. To illustrate how each verse adheres to the iambic meter, we mark unstressed syllables with "x" and stressed syllables with "X".

\section{Dataset}
We opted for the Corpus of Czech Verse~\cite{ccv2015}, curated by the Institute of Czech Literature of the Czech Academy of Sciences.\footnote{\url{https://github.com/versotym/corpusCzechVerse}}
This corpus comprises 1,305 volumes of poetry, each annotated for poetic meters, rhymes, phonetic transcription, word tokenization, lemmatization, and morphological tagging. The annotation is semi-automatic and can thus contain errors; e.g.\ meter annotation has an estimated accuracy of 95.3\%~\cite{ccv2016}. The metadata include information such as the author name, book editors, and the publication years of the book.

\subsection{Dataset Preprocessing}
The utilized corpus lacks direct specification of rhyme schemes, instead providing information on whether two or more verses rhyme or if a verse is non-rhyming. This is carried through the whole poem utilizing numbering system. We transformed this information into standardized rhyme schemes such as \textbf{AABB, AABBCC}, loosing information about between strophe rhyming in progress. But given we only generate individual strophes, this loss is not consequential.
Given that the metadata lacks details about the type of poetry (Lyric, Narrative) or the specific style in which a poem was composed, we inferred that the publication year of the book containing the poem serves as the most indicative feature. However, as Language Models struggle with numerical data and benefit from fine-tuning for improved comprehension~\cite{Spithourakis_2018}, we bucketized the publishing year data into 20-year periods to better categorize poems into distinct styles. Some poems lacked information about their publication year, and for these instances, we introduced the category NaN to encompass such examples.

\subsection{Dataset Makeup}

\begin{figure*}
    \begin{subfigure}{.5\textwidth}
    \includegraphics[width=\textwidth,height=4cm,]{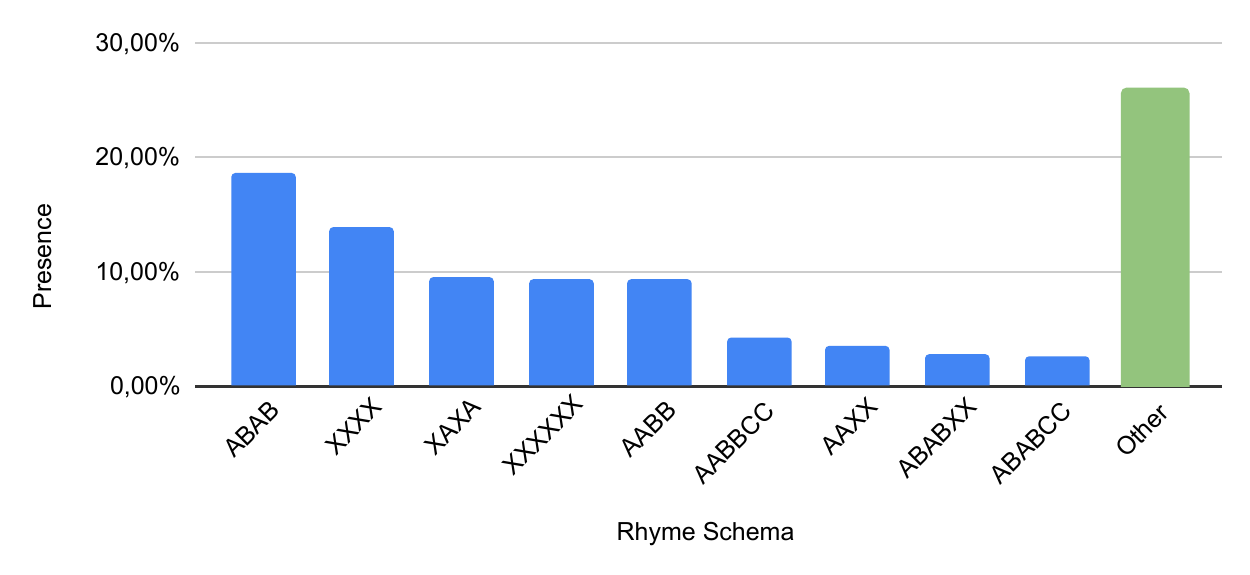}
    \caption{Top 10 rhyme schemes presence}
    \label{fig:rhyme-schema-chart}
    \end{subfigure}%
    \begin{subfigure}{.5\textwidth}
    \includegraphics[width=\textwidth,height=4cm,]{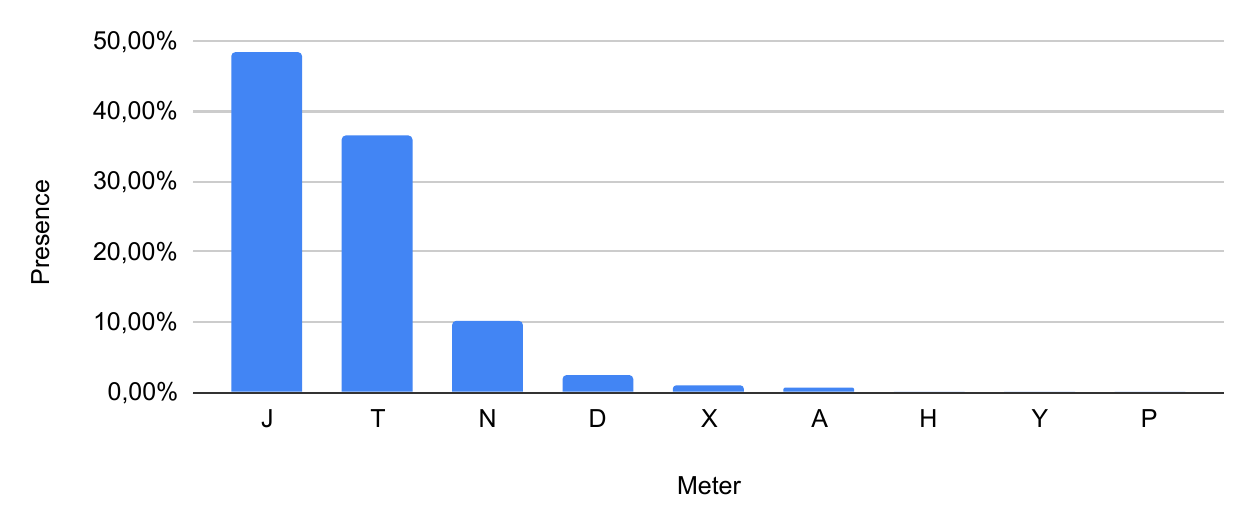}
    \caption{Meter presence}
    \label{fig:meter-chart}
    \end{subfigure}%
    \caption{Rhyme and meter presence}
\end{figure*}

\begin{figure}
    \centering
    \resizebox{\columnwidth}{!}{%
    \includegraphics[width=\textwidth,height=7cm]{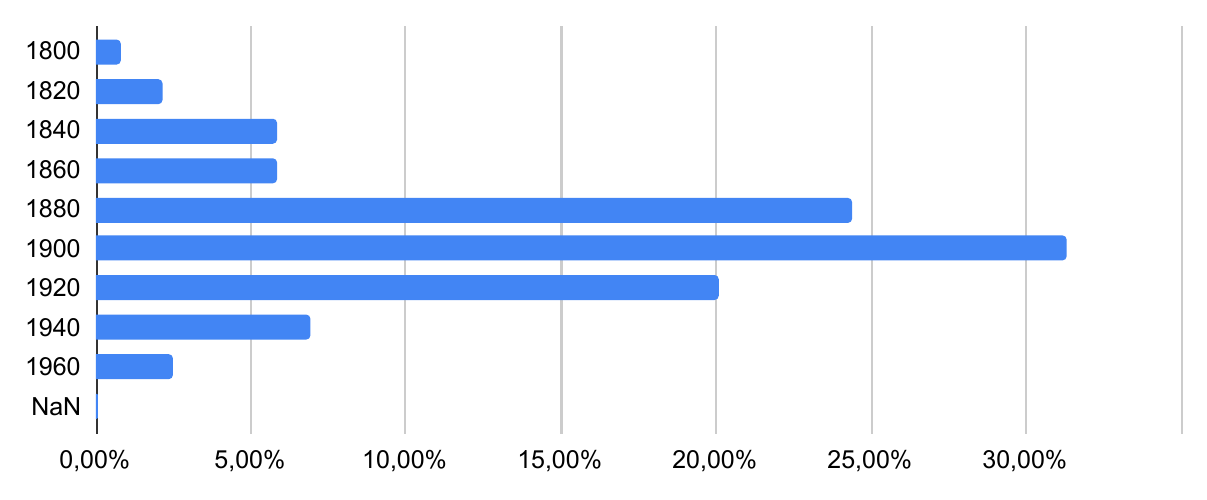}
    }
    \caption{Year regions presence}
    \label{fig:year-chart}
\end{figure}%

To gain a more comprehensive understanding of potential biases in our model, it was crucial to scrutinize the composition of the processed data. The combined corpus encompasses 2,310,917 verses, forming 374,537 strophes, which collectively constitute 66,428 poems.
We split the dataset into a train set (95\%) and a test set (5\%).

\paragraph{Rhyme schemes}
Our processing identified 218 different schemes (primarily due to our leniency towards non-rhyming verses), with a very uneven distribution.
Figure \ref{fig:rhyme-schema-chart} depicts the 10 most frequent rhyme schemes, which together constitute 74\% of the dataset. Conversely, we identified 149 distinct rhyme schemes with a presence below 0.05\% each (fewer than 200 strophes) in our corpus, thus probably constituting noise rather than meaningful patterns that our model could learn from.

\paragraph{Meter}
We observe a modest variety with only 9 distinct types of meter (8 metric and 1 default). However, as illustrated in Figure \ref{fig:meter-chart}, over 85\% of all verses pertain to either iamb (J) or trochee (T), whereas the least frequent meter types (H, Y, P) each individually constitute less than 0.2\% of the data.
Therefore, in the absence of specific instructions, our model is likely to mostly generate J and T verses.

\paragraph{Year of poem publication}
Figure \ref{fig:year-chart} illustrates a more even distribution across all categories than for rhyme schemes and meters. Only NaN exhibits a presence below 0.5\%, while 6 out of the 10 defined regions have a presence exceeding 5\%.

\section{Data Format}
\label{smodelinput}

Standard language modelling is done on the plain text. However, for poetry modelling, previous works have demonstrated strong benefits of explicitly encoding various properties within the text by using annotations via functional tokens interleaved with the actual language tokens.
We therefore explore three variants of specifying strophe and verse parameters.

\paragraph{BASIC} Our initial method, as previously explored in the ByGPT5 article~\cite{belouadi2023bygpt5}, involves adding the rhyme scheme, theme (i.e.\ publishing year), and the most prevalent meter as the first line, while the subsequent lines contain the strophe in plain text; see the example in Figure~\ref{tab:basic-input-schema}.

\begin{figure}[H]
    \begin{tabular}{|l}
    \# ABAB \# 1900 \# J \\
    Tvá loď jde po vysokém moři,\\
    v ně brázdu jako stříbro reje, \\
    svou přídu v modré vlny noří \\
    a bok svůj pěnné do peřeje.
    \end{tabular}
    \caption{Example of a strophe using the BASIC model input format.}
    \label{tab:basic-input-schema}
\end{figure}

\paragraph{VERSE\_PAR} While the initial approach is promising, insights from the GPoet-2 article~\cite{lo2022gpoet2} indicate that relying solely on raw attention may be insufficient, necessitating reverse modeling to achieve rhyming verses. In response to this, we propose the inclusion of a set of verse parameters, \textbf{syllable count} and \textbf{ending hint}, as a prefix to each line, to provide more guidance to the attention mechanism in individual verses.
The ending hint is the Czech poetic clausula according to the rhyming rules of Czech poetry and is the part of the verse that should match its rhytmic partner \cite{czechversebook}.
This modification is reflected in the example in Figure~\ref{tab:verse-input-schema}.

\begin{figure}[H]
    \begin{tabular}{|l}
    \# ABAB \# 1900 \# J \\
    9 \# oři \# Tvá loď jde po vysokém moři,\\
    9 \# eje \# v ně brázdu jako stříbro reje, \\
    9 \# oří \# svou přídu v modré vlny noří \\
    9 \# eje \# a bok svůj pěnné do peřeje.
    \end{tabular}
    \caption{Example of a strophe using the VERSE\_PAR model input format with verse parameters.}
    \label{tab:verse-input-schema}
\end{figure}

\paragraph{METER\_VERSE} Building upon our prior considerations, and given the availability of data for the meter of each individual verse, we recognize the potential value in incorporating meter information for each verse individually instead of the full strophe. This additional input, which can vary between sets of rhyming verses (e.g., from iamb to trochee), provides enhanced guidance to the attention mechanism, particularly in achieving a clear separation of non-rhyming verses. The resulting input scheme is illustrated in Figure~\ref{tab:meter-vesre-input-schema}.

\begin{figure}[H]
    \begin{tabular}{|l}
    \# ABAB \# 1900 \\
    J \# 9 \# oři \# Tvá loď jde po vysokém moři,\\
    J \# 9 \# eje \# v ně brázdu jako stříbro reje, \\
    J \# 9 \# oří \# svou přídu v modré vlny noří \\
    J \# 9 \# eje \# a bok svůj pěnné do peřeje.
    \end{tabular}
    \caption{Example of a strophe using the METER\_VERSE model input format with meter as verse parameter.}
    \label{tab:meter-vesre-input-schema}
\end{figure}

\section{Tokenization}
\label{stokenization}

We recognize tokenization as a critical element in our task,
given our emphasis on formal aspects (rhyming, meter) rather than meaning,
as well as our explicit inclusion of functional tokens specifying desired properties (rhyming, meter, year) interleaved with actual language tokens. We embarked on a series of experiments to address the following objectives:
\begin{itemize}
    \item Distinguish between actual language tokens and functional tokens.
    \item Segment words into tokens that aid in guiding meter and inflection.
    \item Facilitate the swapping of small chunks to encourage fitting the formal requirements and the generation of neologisms.
\end{itemize}

The standard approach in current NLP is subword tokenization, such as BPE~\cite{bpe}.
Given the nature of the Czech language with its reliance on inflection, our focus on formal properties, and the incorporation of neologisms in poetry, particularly for rhyming purposes, we also drew inspiration from approaches involving the separation of words into syllables~\cite{oncevay2020revisiting} or even individual characters~\cite{xue2022byt5}.

Therefore, we experiment with the following four tokenization approaches:
\begin{description}
    \item[BASE] The original tokenizer of the \texttt{
czech-gpt2-oscar} model~\cite{medicalczgpt} which we use.
    \item[OUR] A BPE tokenizer trained on our dataset.
    \item[SYLLABLE] Splitting the text into syllables, using
    the \textit{Sekacek} tool~\cite{sekacek}.\footnote{\url{https://github.com/Gldkslfmsd/sekacek}}
    \item[UNICODE] Splitting the text into individual characters.
\end{description}

The benefit of training a standard BPE tokenizer on our dataset is that it can learn to keep functional annotations as single tokens, as shown in Figure~\ref{fig:token-swap}.\footnote{Of course, this is only effective for annotations that are sufficiently frequent in our dataset.}

\begin{figure}[H]
    \centering
    \begin{tabular}{ll}
    INPUT: &  \# ABAB \# 1900 \\
    BASE: &  [\#] [ AB] [AB] [ \#] [ 1900] \\
    OUR: &  [\#] [ ABAB] [ \#] [ 1900] \\
    SYLL.: & [\#] [ ABAB] [ \#] [ 1900] \\
    UNIC.: & [\#][ ][A][B][A][B][ ][\#][ ][1][9][0][0]
    \end{tabular}
    \caption{Tokenization of strophe parameters.
    }
    \label{fig:token-swap}
\end{figure}

Obviously, SYLLABLE and UNICODE encode sequences into larger amounts of shorter tokens; see Figure~\ref{fig:our-tokenizers}.
This allows the model to make fine generation decisions with a higher granularity, so that it can better fit the prescribed formal properties (meter, rhyme).
It also makes production of neologisms easier.
However, as mentioned by ByGPT5 article~\cite{belouadi2023bygpt5}, the time required for model training and inference increases accordingly.

\begin{figure}[H]
    \centering
    \begin{tabular}{ll}
    INPUT: & a v duchu \\
    BASE: & [a] [ v] [ duchu] \\
    OUR: & [a] [ v] [ duchu] \\
    SYLLABLE: & [a] [ v] [ duch] [u] \\
    UNICODE: & [a] [ ] [ ] [v] [ ] [d] [u] [c] [h] [u]\\
    \end{tabular} 
    \caption{Tokenization of verse text.}
    \label{fig:our-tokenizers}
\end{figure}

\section{Training the Models}

As our base model, we have selected \texttt{czech-gpt2-oscar} by Chaloupský~\shortcite{medicalczgpt},\footnote{\url{https://huggingface.co/lchaloupsky/czech-gpt2-oscar}} a GPT-2-small model~\cite{Radford2019LanguageMA} trained on the Czech part of the OSCAR dataset~\cite{Ortiz_Su_rez_2020}.
This selection was made based on the low availability of Czech models.

We then fine-tune the model on our dataset, using one of the three data formats (see \nameref{smodelinput} Section) and one of the four tokenizers (see \nameref{stokenization} Section).

We explore two different approaches of training the model for the selected data format.
We either simply train the model using only the selected data format, or we first pre-train the model using the METER\_VERSE format, and then fine-tune it using the BASIC or VERSE\_PAR format;
the motivation for this approach is that each of these formats can be regarded as a subset of the METER\_VERSE format.

For our loss computation, we employ the conventional Cross Entropy Loss, with our input serving as labels as well. Since our model is GPT-based, we avoid using input masking because GPT-2 is trained more effectively through next word prediction, which is the preferred training method.

For using our custom tokenizers, we have followed the model recycling approach~\cite{recycling}, which utilizes overlap in current and target vocabularies to jump-start the model by keeping large parts of the embedding matrix.

\begin{table}
    \centering
    \begin{tabular}{lr}
    \toprule
     \textbf{Tokenizer} & \textbf{Model parameters}  \\
    \midrule
    BASE & 137M \\
    OUR & 137M \\
    SYLLABLE & 105M \\
    UNICODE & 86M \\
    \bottomrule
    \end{tabular}
    \caption{Sizes of the fine-tuned models, depending on the tokenization approach.
    }
    \label{tab:model-sizes}
\end{table}

The sizes of the resulting fine-tuned models can be seen in Table~\ref{tab:model-sizes}. As SYLLABLE and UNICODE tokenizers have smaller vocabularies, the resulting models are smaller;
on the other hand, the data format has no effect on the model size.

\section{Text Generation}

To further enhance the model's proficiency in adhering to strophe and verse parameters at inference,
we propose an alternative approach to the standard text generation method.

\paragraph{Basic Decoding} The prompt consists of the first line which specifies the strophe parameters. Then, generation proceeds token by token until the end-of-sequence token is generated.

\paragraph{Forced~Generation}
This iterative method involves examining an already accepted rhyme scheme and compelling verse parameters for lines intended to rhyme. After generating each verse, the generation process stops, and if the next verse to be generated should rhyme with an already generated verse, then the verse parameters are copied (forced) as the prefix for the next line before resuming the generation process,
as illustrated in Figure~\ref{tab:forced-generation}.
More formally, if the model has already generated meter \textit{X}, syllable count \textit{Y} and ending hint \textit{Z} as annotations for a verse connected to character \textbf{A} in the rhyme scheme, all other verses linked to character \textbf{A} will be prompted with verse parameters \textit{X \# Y \# Z \#}.
Obviously, this approach is only applicable for VERSE\_PAR and METER\_VERSE input formats.

\begin{figure}[H]
    \begin{tabular}{|l}
    \# AABB \# 1900  \\
    \textbf{T \# 8 \# ání \#} A když přijde z nenadání, \\
    \underline{T \# 8 \# ání \#}  ... 
    \end{tabular}
    \caption{\textit{Forced~generation}. According to the AABB rhyme scheme, the second verse should rhyme with the first verse. Thus, after generating the first verse, the verse parameters for the second verse (\underline{underlined}) are forced, i.e.\ copied from the first verse (\textbf{in bold}).}
    \label{tab:forced-generation}
\end{figure}

\section{Validators}

Comprehensive automated quality evaluation of text generation is hard. In our setting, we have decided to focus on a narrower subtask, mostly evaluating formal quality of the generated poetry.
Other approaches exist~\cite{rhymetagger}, but given the large annotated dataset at our disposal, we can train validator models directly on the dataset.
Specifically, we train classifiers that label strophes with the \textbf{rhyme scheme}, \textbf{meter}, and \textbf{year}. We can then simply evaluate whether the predicted value matches the value specified on the input.

The general approach we take is to train a softmax classifier attached to the class token representation in a masked language model; we use either RoBERTa~\cite{liu2019roberta}, or its Czech version, RobeCzech~\cite{Straka_2021}.

\subsection{Validator Input Preprocessing}
\paragraph{Syllabification} As syllables are useful text units when concerned with formal properties of poetry, we again experiment with splitting the input into syllables before feeding it into the RoBERTa/RobeCzech model. This approach could simplify the tasks of rhyme and meter validators, as they no longer need to guess word partitioning. However, the effectiveness of syllabification for the year validator is uncertain. Understanding themes requires both grasping the employed metrical and rhyming structures, as well as discerning the semantic meaning, where syllabification could cause a partial disruption.

\paragraph{Contextualization} While the meter is a fundamental verse parameter, poets usually do not follow it exactly. The resulting sequence of weak and strong syllables, called rhythm, could thereafter be argued to fulfill multiple meters. To prevents this, the context of the whole verse is taken into account, as it is uncommon to employ multiple meters in a classical rhythmic strophe.

\subsection{Validators Accuracies}

Using the train and test parts of the dataset, we train and evaluate validators for rhyme scheme prediction (Table~\ref{tab:rhyme_acc}), meter prediction (Table~\ref{tab:meter_acc}) and publishing year prediction (Table~\ref{tab:year_acc}).
We also report the \textbf{Baseline} as the most common class,
and for meter, we have included an \textbf{Upper bound} based on the accuracy of the semi-automatic annotation in the dataset~\cite{ccv2016}.
We measured statistical significance of the results using Random Permutation Test \cite{permutationtest} with 100 repetitions on the test set.

\begin{table}
    \centering
    \begin{tabular}{lcc}
    \toprule
    \textbf{Base model} & \textbf{Input type}  & \textbf{Accuracy}  \\
    \midrule
    robeczech-base & Syllable  & 94.68 \%  \\
    robeczech-base & Raw  & 48.06 \%  \\
    \midrule
    roberta-base & Syllable  & 96.62 \%  \\
    roberta-base & Raw  & \textbf{96.96 \%} \\
    \midrule
    \textit{Baseline} &  & \textit{18.65 \%} \\
    \bottomrule
    \end{tabular}
    \caption{Rhyme scheme prediction validator.}
    \label{tab:rhyme_acc}
\end{table}
\begin{table}
    \centering
    \begin{tabular}{lcc}
    \toprule
    \textbf{Base model} & \textbf{Input type}  & \textbf{Accuracy}  \\
    \midrule
    robeczech-base & Syllable & 89.90  \%  \\
    robeczech-base & Raw & 90.03 \%  \\
    robeczech-base & Contextual & \textbf{94.94} \%  \\
    \midrule
    roberta-base & Syllable & 89.12 \%  \\
    roberta-base & Raw & 89.28 \%  \\
    roberta-base & Contextual & \textbf{94.34} \%  \\
    \midrule
    \textit{Baseline} &  & \textit{48.52 \%} \\
    \textit{Upper bound} &  & \textit{95.30 \%} \\
    \bottomrule
    \end{tabular}
    \caption{Meter prediction validator. Accuracy of meter prediction for each verse.}
    \label{tab:meter_acc}
\end{table}
\begin{table}
    \centering
    \begin{tabular}{lcc}
    \toprule
    \textbf{Base model} & \textbf{Input type}  & \textbf{Accuracy}  \\
    \midrule
    robeczech-base & Syllable  & 42.55 \% \\
    robeczech-base & Raw  & \textbf{47.45 \%} \\
    \midrule
    roberta-base & Syllable  & 40.97 \% \\
    roberta-base & Raw   & 43.15 \%  \\
    \midrule
    \textit{Baseline} &  & \textit{31.33 \%} \\
    \bottomrule
    \end{tabular}
    \caption{Year of publishing prediction validator.}
    \label{tab:year_acc}
\end{table}

\paragraph{Syllabification} Pre-splitting the input into syllables aids the RobeCzech validator in classifying rhyme scheme, but was significantly harmful for the year classification.
Year classification deterioration aligns with our expectations, as the year of publishing is more closely tied to the subject of the poem, a facet disrupted by the syllabification process.

\paragraph{Contextualization} Inclusion of the whole strophe in meter prediction showed great improvement, with the resulting validator almost achieving the set upper bound.
Adding syllabification to contextualization had negligible effect on the accuracy, we therefore report only accuracies with non-syllabified (raw) text. The resulting model utilizing contextualization was significantly better than counterparts that did not implement this preprocessing step.

\paragraph{Rhyme scheme and meter prediction} The validators on syllabified input achieve very high accuracies, reaching or approaching the maximum accuracies achievable on the dataset, as the semi-automated annotation of the dataset is not perfect and contains errors.
The accuracies of RobeCzech and RoBERTa are close;
RoBERTa is significantly better than RobeCzech for rhyme prediction, but the difference is not statistically significant for meter prediction.

\paragraph{Year prediction} Using RobeCzech leads to significantly higher accuracies than using RoBERTa. We believe this is because this task also requires understanding the semantics of the text, whereas the other tasks focus on the formal properties of the text, and thus the model pre-trained on Czech data has a notable advantage. Still, all the accuracies on this task are rather low, and we do not deem them sufficient for using this validator to reliably evaluate the results of poetry generation.

\paragraph{Token granularity} In the context of rhyme scheme,
we have observed that the effect of syllabification
is less pronounced for RoBERTa than for RobeCzech.
We posit that this is because RoBERTa is not pre-trained on Czech texts and thus its subword tokenization needs to split the text into shorter tokens to represent Czech words.

\begin{table}[H]
    \centering
    \begin{tabular}{lr}
    \toprule
    \textbf{Tokenizer} & \textbf{ Chars per token} \\
    \midrule
    roberta-base & 1.50 \\
    robeczech-base & 2.72 \\
    \bottomrule
    \end{tabular}
    \caption{Tokenizer influence on token granularity}
    \label{tab:val-tokenizers}
\end{table}

\label{validators-accs}

We evaluated the model tokenizers by analyzing 10,000 verses and calculating the average number of characters per token.
As showcased in Table \ref{tab:val-tokenizers}, RoBERTa already tokenizes the text more granularly, resulting in further syllabification having a weaker effect than in the case of RobeCzech.

\section{Model Validation}
\label{seval}

\begin{table}[b]
    \centering

    \resizebox{\columnwidth}{!}{%
    \begin{tabular}{lccc}
    \toprule
    \textbf{Data format} & \textbf{Pre-train} & \textbf{Rhyme acc} & \textbf{Meter acc} \\
    \midrule
    BASIC & False & 49.6 \% & 94.4 \% \\
    BASIC & True & 41.1 \% & 92.2 \% \\
    \midrule
    VERSE\_PAR & False & \textbf{ 89.8 \%} & 94.4 \% \\
    VERSE\_PAR & True & \textbf{88.3 \%} & 94.6 \% \\
    \midrule
    METER\_VERSE &  & \textbf{86.8 \%} & 94.6 \% \\
    \bottomrule
    \end{tabular}
    }
    \caption{Influence of Data Format on accuracy.}
    \label{tab:input-schema-influence}
    
\end{table}

\begin{table*}[t]
\centering
    \begin{tabular}{lcccccc}
    \toprule
     \textbf{Tokenizer} & \textbf{Generation} & \textbf{Num syl} & \textbf{End acc} & \textbf{Unique} & \textbf{ Rhyme acc}  & \textbf{Meter acc}\\
     \midrule
     BASE & Basic  & 91.8 \% & 94.6 \% & 85.7 \% & 86.5 \%  & 94.5 \%  \\
     BASE & Forced  & 92.3 \% & 95.0 \% & 84.9 \% & 86.9 \%  &  93.8 \%  \\
     \midrule
     OUR & Basic  & 91.0 \% & 95.4 \% & 83.3 \% & 80.6 \% &  94.6 \%  \\
     OUR & Forced  & 90.6 \% & 95.4 \% & 83.6 \% & 80.6 \% & 94.8 \%  \\
     \midrule
     SYLLABLE & Basic  & 94.4 \% & 95.2 \% & 85.6 \% & 88.7 \% &  94.6 \%  \\
     SYLLABLE & Forced  & 93.8 \% & 95.3 \% & 85.0 \% & 87.7 \% &  94.2 \% \\
     \midrule
     UNICODE & Basic  & 91.1 \% & 91.2 \% & 89.7 \% & 73.8 \% & 93.5 \%  \\
     UNICODE & Forced  & \textbf{97.8 \%} & \textbf{98.0 \%} & \textbf{88.8 \%} & \textbf{94.0 \%} &  94.0 \%  \\
     \midrule
     \textit{Target} &  & \textit{100 \%} & \textit{100 \%} & \textit{87.9 \%} & \textit{100 \%} & \textit{100 \%} \\
     \bottomrule
    \end{tabular}%
\caption{Validation results for the final models.}
\label{tab:final-acc}
\end{table*}

Through our validators, we can evaluate the poetry generation model's adherence to the rhyme scheme and meter. In addition to these metrics, we also assess conformity to the number of syllables and the ending hint for each verse as generated (or forced) in the prefix annotation at the start of the line.
We also measure the uniqueness of the generated syllables as an indicator for non-repetitiveness.
Altogether, we compute these characteristics:
\begin{description}
    \item[Num syl] Proportion of verses with number of syllables matching the prefix annotation.
    \item[End acc] Proportion of verses with ending hint matching the prefix annotation.
    \item[Unique] Ratio of unique syllables among all syllables in the strophe; the optimal value here is not 100\%, but rather the value observed on the true data in the dataset (87.9\%).
    \item[Rhyme acc] Proportion of strophes with rhyme scheme matching the first line annotation.
    \item[Meter acc] Proportion of strophes with the meter of all verses matching the annotation.
\end{description}

We use the annotations of the strophes in the test part of our dataset as inputs (as in Figure~\ref{tab:generation-prompt}), and evaluate the generated outputs (now disregarding the actual texts of the strophes in the test dataset).

\begin{figure}[H]
    \begin{tabular}{|l}
    \# AXAX \# 1880 \\
    J \# ... 
    \end{tabular}
    \caption{Example of an input prompt using METER\_VERSE format.}
    \label{tab:generation-prompt}
\end{figure}

\subsection{Influence of Data Format}

We first evaluate the effect of the data format (see \nameref{smodelinput} Section), while using the BASE tokenizer and Basic text generation.

The model was either trained using only the selected data format for 16 epochs, or it was first pretrained using METER\_VERSE format for 16 epochs and then fine-tuned for further 4 epochs using the selected format.

Table \ref{tab:input-schema-influence} demonstrates that incorporating the individual verse parameters using either VERSE\_PAR or METER\_VERSE format significantly contributes to the model performance, particularly in terms of adhering to the rhyme scheme. The inclusion of more detailed meter parameters in METER\_VERSE scheme slightly enhances the models meter accuracy. Furthermore, the performance with  VERSE\_PAR format does not change when the model is first pre-trained using the METER\_VERSE format. Notable decrease was observed using the BASIC format. We posit that it is the result of large difference between BASIC and METER\_VERSE Input.

\subsection{Final Validation}

Finally, we train four models, exploring all the presented tokenizers (BASE, OUR, SYLLABLE, UNICODE), using the METER\_VERSE data format, and training for 16 epochs.
We generate strophes using either \textit{Basic~decoding} or \textit{Forced~generation}.

As shown in Table~\ref{tab:final-acc}, the best results are obtained by using the UNICODE tokenizer and \textit{Forced~generation},  surpassing the other setups. This underscores the viability of character-level large language models, particularly in morphological and phonetic tasks.\footnote{On the other hand, we expect that character-level processing has a negative effect on semantic quality of the generated texts; however, we have not evaluated semantic quality in this paper and leave this for future work.}
For meter accuracy, all of the setups showed high performance, with statistically insignificant differences between the individual setups.

\subsection{Validation Results Analysis}

\paragraph{Forced~Generation} Our proposed approach to generation significantly uplifted the performance of UNICODE tokenizer model, making it the best model in most categories. We posit that the improvements in rhyme scheme accuracy can be attributed to the fact that \textit{Forced~generation} constrains the model to generate matching verses with the same ending hint, which plays role in rhyming. This constraint is also the reason behind the usual decrease in meter accuracy and unique syllables ratio. The enforced ending hint is not unique, and it compels the model to generate proper meter inclusive of it, which, especially with single-syllable unstressed words, can pose a challenge.

\paragraph{BASE, SYLLABLE, UNICODE tokenizer} Both BASE and SYLLABLE tokenizer performed well, with their rhyme scheme accuracies significantly better than model utilizing OUR tokenizer. The difference between BASE and SYLLABLE was not statistically significant in any metric. UNICODE tokenizer model was significantly the worst when \textit{Forced~generation} was not utilized. On the contrary, with the usage of \textit{Forced~generation} it was significantly the best model in Number of syllables, Ending hint accuracy and Rhyme scheme accuracy.

\paragraph{OUR tokenizer} The performance of OUR tokenizer was the least satisfactory among the considered options with the resulting rhyme scheme accuracy being significantly worse that all remaining models. We contend that this can be attributed to the fact that OUR tokenizer was trained solely on poetry data, comprising only 2 GB in size. The resulting number of characters per token is excessively large, rendering it less efficient for poetry generation. Unlike SYLLABLE or UNICODE tokenizer, OUR tokenizer lacks the capability for syllable or character substitution. To substantiate this observation, we conducted the same analysis as for validator tokenizers (see \nameref{validators-accs} Section, Table~\ref{tab:val-tokenizers}).
In Table~\ref{tab:model-tokenizers}, we can observe that OUR tokenizer encodes 3.77 characters per token, which is the highest value among all tokenizers. This characteristic diminishes flexibility, restricting words to be represented by only 1 token.

\begin{table}
    \centering
    \begin{tabular}{lr}
    \toprule
    \textbf{Tokenizer} & \textbf{Chars per token} \\
    \midrule
    BASE & 3.37 \\
    OUR & 3.77 \\
    SYLLABLE & 2.43 \\
    UNICODE & 1.00 \\
    \bottomrule
    \end{tabular}
    \caption{Tokenizer influence on token granularity}
    \label{tab:model-tokenizers}
\end{table}

\subsection{Year Accuracy}

Driven by curiosity, we also employed our validator to assess the probable publishing year accuracy, which is our proxy for poetic style; keeping in mind that this validator is highly unreliable as its accuracy is rather low. Our hypothesis was grounded in the belief that OUR tokenizer, with its capacity to tokenize entire words in a single token, might excel in tasks oriented more towards semantic meaning.

\begin{table}
    \centering
    \begin{tabular}{lr}
    \toprule
    \textbf{Tokenizer} & \textbf{Year accuracy} \\
    \midrule
     BASE & 41.1 \% \\
     OUR & 41.2 \% \\
     SYLLABLE & 41.1 \%  \\
     UNICODE & 38.6 \% \\
     \bottomrule
    \end{tabular}
    \caption{Year accuracy as reported by the validator model. For each tokenizer, we report the best result observed among all investigated configurations. Note that the year validator is highly unreliable.}
    \label{tab:year-accs}
\end{table}

Given the high unreliability of the used validator, the results showcased in Table \ref{tab:year-accs} do not provide much of a meaning. The UNICODE tokenizer model is the only one, that seems to showcases lower semantic understanding of the strophes. This was expected as UNICODE tokenizer trades flexibility as evident by Table \ref{tab:model-tokenizers} and collaborated by Table \ref{tab:final-acc}.

\section{Conclusion}
In this work, we proposed and implemented a novel comprehensive approach to poetic strophe generation, focusing on formal qualities of poetry. We trained and evaluated our models using a corpus of Czech poetry.

Our results reveal that enriching the plain text with interleaved explicit annotations (number of syllables, verse ending hint) significantly helps to better guide the model.
We have also demonstrated a superior rhyming accuracy of character tokenization compared to standard subword tokenization methods when specific constraints are followed by \textit{Forced~generation}.

In future work, we want to expand our generation to full poems with strophes that are thematically and schematically connected.

\section{Ethical Considerations}

A topic of active discussion is whether it is ethical (or even legal) to use various kinds of data for training large language models, e.g.\ without explicit consents of the data authors.
In our work, we train the language model on a dataset composed exclusively of poems in the public domain (due to the authors having died more than 70 years ago), which we consider to be non-problematic.

The base GPT-2 model, which we further fine-tune on that dataset, was trained on various kinds of data, including potentially problematic data. However, our approach can be in principle applied to any base model; thus, if there is ever a consensus that it is not ethical to use this base model, our approach can be repeated and reevaluated using any other base model.

It is becoming the norm (and may be soon required by laws, such the EU AI Act) to label automatically generated works as such, e.g.\ to avoid unintentional spreading of misinformation. To this end, we make sure to always label all our generated poems as automatically generated.

\section{Limitations}
As any transformer model, our solution grapples with substantial computational complexity~\cite{vaswani2017attention}, necessitating the use of powerful GPUs (A40 40GB, A100 40GB, H100 80GB) for effective training.

An inherent challenge arises from the use of multiple tokenization techniques, potentially impacting the scalability of next strophe generation. Notably, the UNICODE tokenizer struggles to retain context across two verses, posing a risk of losing crucial information.

Another issue stems from data distributions, as illustrated in Figures \ref{fig:rhyme-schema-chart} and \ref{fig:meter-chart}. If not prompted appropriately, the model defaults to a rhyme scheme of ABAB and a meter of iamb. This default behavior is problematic, particularly considering that the model is likely incapable of generating most of the 218 rhyme schemes appearing in the dataset. Regarding meter, only iamb, trochee, and free-verse are reliably generated, with the remaining 6 typically defaulting to iamb.

With our inability to observe if year of publishing is followed (Table \ref{tab:year_acc}), it remains uncertain whether the model gains any meaningful information from this parameter.

Lastly, we intentionally disregarded the meaning in poems and notably simplified our measures around strophe uniqueness. Generated verses tend to repeat entire words and syllables to create the illusion of rhyming, whereas a more preferable approach would involve generating syllables with close phonetics.


\section{Acknowledgments}
The work has been partially supported by the EduPo grant (TQ01000153 Generating Czech poetry in an educative and multimedia environment), which is co-financed from the state budget by the Technology agency of the Czech Republic under the SIGMA DC3 Programme.
Computational resources were provided by the e-INFRA CZ project (ID:90254), supported by the Ministry of Education, Youth and Sports of the Czech Republic.
The work described herein has also been using data, tools and services provided by the LINDAT/CLARIAH-CZ Research Infrastructure (https://lindat.cz), supported by the Ministry of Education, Youth and Sports of the Czech Republic (Project No. LM2023062).

\bibliographystyle{iccc}
\bibliography{iccc}

\end{document}